\documentclass{article}
\usepackage{ijcai26}

\usepackage{graphicx}

\usepackage{times}
\usepackage{soul}
\usepackage{url}
\usepackage[utf8]{inputenc}
\usepackage[small]{caption}
\usepackage{graphicx}
\usepackage{amsmath}
\usepackage{amsthm}
\usepackage{mathtools}
\usepackage{booktabs}
\usepackage{algorithm}
\usepackage{algorithmic}
\usepackage[switch]{lineno}

\usepackage{amssymb}
\usepackage{enumitem}
\usepackage{tabularx}
\usepackage{array}
\usepackage{multirow}
\usepackage{dsfont}
\usepackage{xcolor}
\usepackage{pifont} 
\usepackage[hidelinks]{hyperref}
\usepackage[printonlyused]{acronym}

\newcolumntype{R}{>{\raggedleft\arraybackslash}X}
\newcolumntype{L}{>{\raggedright\arraybackslash}X}
\newcolumntype{C}{>{\centering\arraybackslash}X}

\definecolor{myred}{HTML}{DC3912} 
\definecolor{mygreen}{HTML}{109618}

\newcommand{\xmark}{\ding{55}}
\newcommand{\cmark}{\ding{51}}

\acrodef{ood}[OOD]{out-of-distribution}
\acrodef{id}[ID]{in-distribution}
\acrodef{fpr95}[FPR@95]{false positive rate at 95\% true positive rate}
\acrodef{auroc}[AUROC]{area under the receiver operating characteristic curve}
\acrodef{oe}[OE]{Outlier Exposure}
\acrodef{msp}[MSP]{maximum softmax probability}
\acrodef{ce}[CE]{cross-entropy}
\acrodef{sgd}[SGD]{stochastic gradient descent}
\acrodef{fft}[FFT]{fast Fourier transform}
\acrodef{rfft}[rFFT]{real fast Fourier transform}
\acrodef{knn}[$k$-NN]{$k$-nearest neighbors}
\acrodef{maha}[Maha++]{Mahalanobis++}
\acrodef{vmf}[vMF]{von Mises--Fisher}
\acrodef{mlp}[MLP]{multilayer perceptron}
\acrodef{har}[HAR]{human activity recognition}
\acrodef{ecg}[ECG]{electrocardiogram}
\acrodef{eeg}[EEG]{electroencephalogram}
\acrodef{eog}[EOG]{electrooculogram}

\title{Learning Hyperspherical Time--Frequency Representations\\for Time-Series Out-of-Distribution Detection}

\author{
Willian T. Lunardi$^{1}$\thanks{Equal contribution. $^\dagger$Corresponding author.}$^\dagger$,
Samridha Shrestha$^{1}$\footnotemark[1]
\textnormal{and}
Martin Andreoni$^{2}$\\
\affiliations
$^1$Technology Innovation Institute, Abu Dhabi, United Arab Emirates\\
$^2$Khalifa University, Abu Dhabi, United Arab Emirates\\
\emails
\{willian.lunardi, samridha.shrestha\}@tii.ae,
martin.andreoni@ku.ac.ae
}

\begin{document}

\maketitle

\begin{abstract}
     \Ac{ood} detection for time-series data remains comparatively underexplored compared to vision and language, with a limited principled understanding of how supervised time-series representations can be leveraged for reliable detection under distributional shifts. This work formulates time-series \ac{ood} detection as representation learning with hyperspherical embeddings, where class-conditional structure is induced by a \ac{vmf} likelihood--based objective on the unit sphere.
    The learned representation combines time- and frequency-domain views of the input signal via domain-specific encoders, integrating them into a joint embedding space for \ac{ood} detection. Detection uses distance-based scores over the learned embeddings, including \ac{knn} and Mahalanobis scores. We evaluate the approach at scale on the complete UCR and UEA time-series archives under a cross-dataset protocol.
    Empirical results show consistent improvements under both \ac{knn} and Mahalanobis scoring over strong contrastive-learning and post-hoc baselines in the same setting. Code is available at \url{https://github.com/tiiuae/hypertf-time-series-ood}.
\end{abstract}

\acresetall

\section{Introduction}

\Ac{ood} detection constitutes a core requirement for dependable machine learning systems, particularly in regimes where deployment-time failures incur disproportionate risk. In such regimes, predictive models are routinely exposed to input distributions that deviate from those observed during training, rendering the ability to identify samples lying outside the intended domain of generalization coequal in importance to achieving strong \ac{id} performance. Although \ac{ood} detection has been extensively investigated in modalities such as images and text, its formulation for time-series data remains comparatively underdeveloped~\cite{lu2024diversify}. This gap is nontrivial: time-series observations exhibit structured temporal dependencies, frequency-domain characteristics, and measurement processes tightly coupled to sensing hardware, acquisition protocols, and preprocessing pipelines, thereby inducing distributional shifts that are qualitatively distinct from those encountered in exchangeable or spatially indexed data.

Existing approaches to \ac{ood} detection are typically grouped into two broad categories: post-hoc scoring methods, which estimate OODness from classifier outputs or intermediate representations obtained under standard supervised training, and training-time methods, which modify the learning objective to improve separation between in- and \ac{ood} samples. Although both categories perform well in many vision-focused benchmarks~\cite{yang2022openood,zhang2023openood}, recent results indicate that their effectiveness often degrades on multivariate time-series data~\cite{gungor2025ts}. A key reason is that distribution shift in time-series settings is rarely purely semantic, but instead reflects combined changes in the underlying process and measurement pipeline, leading to systematic differences in temporal or spectral structure caused by sensing hardware, acquisition protocols, population differences, or preprocessing choices~\cite{baek2024unexplored}. This issue is compounded by evaluation practices adopted from open set recognition, which commonly rely on intra-dataset, label-disjoint splits that assume identical sensing conditions for \ac{id} and \ac{ood} samples. Such protocols underrepresent the cross-environment shifts encountered in practice and, in datasets with few classes, can blur the distinction between open-set recognition (OSR) and \ac{ood} detection.

Motivated by these observations, we approach time-series \ac{ood} detection from a representation learning perspective that explicitly accounts for both semantic structure and geometric organization in embedding space. In particular, we consider hyperspherical embeddings, in which class-conditional structure is induced on the unit sphere through objectives motivated by directional statistical models. Within this framework, distance-based \ac{ood} scoring arises naturally from the geometry of the learned representations. To capture complementary aspects of time-series data, the proposed method integrates time-domain and frequency-domain views through domain-specific encoders, enabling the model to encode both temporal dynamics and spectral characteristics while preserving a unified embedding geometry suitable for \ac{ood} detection. The contributions of this work are summarized as follows:
\begin{itemize}
    \item We propose HyperTF, a hyperspherical representation learning framework for time-series \ac{ood} detection that induces class-conditional structure on the unit sphere via a \ac{vmf} likelihood formulation.
    \item We introduce a dual-view time--frequency architecture that learns aligned representations from temporal and spectral views, coupled through a shared hyperspherical decision geometry and a cross-domain consistency regularizer.
    \item Extensive experiments show that HyperTF improves embedding-level \ac{ood} separability over strong contrastive and post-hoc baselines, particularly under \ac{maha} scoring, and that auxiliary outliers (optionally with MixOE) further enhance detection in limited-coverage regimes.
\end{itemize}

The remainder of the paper is organized as follows. Section~\ref{sec:related_work} reviews related work on \ac{ood} detection in general and time-series settings in particular. Section~\ref{sec:problem-definition} defines the considered problem and introduces the required hyperspherical preliminaries. Section~\ref{sec:methodology} presents the HyperTF framework, while Section~\ref{sec:experiments} reports the experimental results and ablation studies. Finally, Section~\ref{sec:conclusion} concludes the paper.

\section{Related Work}\label{sec:related_work}

\paragraph{Post-hoc confidence scoring.}
Early \ac{ood} detection work primarily focuses on post-hoc confidence estimation, where models are trained normally and \ac{ood} signals are derived at inference via training-agnostic scores. The standard baseline is \ac{msp} \cite{hendrycks2017msp}, with subsequent methods improving logit calibration or confidence modeling, including Energy-based scoring \cite{liu2020energy}, MaxLogits \cite{hendrycks2019scaling}, GEN \cite{liu2023gen}, and LogicOOD \cite{kirchheim2024out}. Distance-based approaches measure deviations from the \ac{id} feature manifold: parametric methods such as the Mahalanobis detector \cite{lee2018simple_maha}, \ac{maha} \cite{mueller2025mahalanobis++}, and SSD \cite{sehwag2021ssd} assume class-conditional Gaussian structure, while non-parametric methods like \ac{knn} \cite{sun2022knn} and NNGuide \cite{park2023nearest} leverage local neighborhood geometry. Other approaches exploit model internals, including gradient- and feature-based methods such as GradNorm \cite{huang2021importance} and GAIA \cite{chen2023gaia}, as well as activation-based techniques like ODIN \cite{liang2017enhancing}, ReAct \cite{sun2021react}, ViM \cite{wang2022vim}, and ASH \cite{djurisic2022extremely}. From a density-modeling perspective, feature-space methods such as GEN \cite{liu2023gen} and ConjNorm \cite{peng2024conjnorm} have also been explored.

\paragraph{Training-time regularization.}
A complementary line of \ac{ood} detection research alters training objectives or embedding geometry to better separate in- and \ac{ood} samples. Early work embeds \ac{ood} awareness directly into representation learning without auxiliary data, often via reconstruction-based objectives, where methods such as MoodCat \cite{yang2022out}, MOOD \cite{li2023rethinking}, and DiffGuard \cite{gao2023diffguard} identify \ac{ood} inputs through reconstruction errors. Other approaches calibrate features or logits using probabilistic, normalization, or density-based regularization, including HVCM \cite{li2023hierarchical}, DDR \cite{huang2022density}, LogitNorm \cite{wei2022mitigating}, and DML \cite{zhang2023decoupling}. More recent work explicitly shapes embedding spaces with contrastive or metric learning, building on NT-Xent \cite{chen2020simple} and SupCon \cite{khosla2020supervised}, and extending to \ac{ood}-aware variants such as \citeauthor{winkens2020contrastive}~(\citeyear{winkens2020contrastive}), CSI \cite{tack2020csi}, SSD+ \cite{sehwag2021ssd}, KNN+ \cite{sun2022knn}, and FIRM \cite{lunardi2025contrastive}. Complementary prototype-based and hyperspherical methods impose class-conditional geometry to enhance \ac{ood} separation, including Step \cite{zhou2021step}, SIREN \cite{du2022siren}, CIDER \cite{ming2023cider}, PALM \cite{lu2024learning}, ReweightOOD \cite{regmi2024reweightood}, and PFS \cite{wu2024pursuing}. \citeauthor{burapacheepyour}~(\citeyear{burapacheepyour}) further show that unit-norm classifiers induce a \ac{vmf} mixture density whose log-likelihood supports principled \ac{ood} detection.

\paragraph{Auxiliary outliers.}
Another line of work incorporates auxiliary outlier data during training to enhance \ac{ood} detection by shaping model behavior near or beyond the \ac{id} boundary, typically promoting uniform or low-confidence predictions for \ac{ood} inputs. Such auxiliary-outlier-based methods depend on external datasets or sampling strategies to provide explicit \ac{ood} supervision, including \ac{oe}~\cite{hendrycks2019oe}, MCD~\cite{yu2019unsupervised}, UDG~\cite{yang2021semantically}, ATOM~\cite{chen2021atom}, and POEM~\cite{ming2022poem}. Despite their effectiveness, these methods are sensitive to the quality and diversity of the selected outliers, which has motivated subsequent approaches such as MixOE~\cite{zhang2023mixture}, DivOE~\cite{zhu2023diversified}, and DiverseMix~\cite{yao2024out}. These later methods improve boundary coverage through mixup-based augmentation or learnable extrapolation strategies.

\paragraph{\Ac{ood} detection for time series.}
\Ac{ood} detection for time-series data remains largely underexplored compared to vision and language, despite its importance in industrial, healthcare, and other safety-critical settings. This is partly due to the field's historical focus on internal structural anomalies rather than distributional shifts \cite{lai2021revisiting}. Only recently has a benchmark been proposed: to our knowledge, \citeauthor{gungor2025ts}~(\citeyear{gungor2025ts}) introduce the first modality-agnostic evaluation on multivariate time series, but it is limited to intra-dataset semantic class splits, lacks auxiliary outlier data, and does not consider modality-aware or cross-dataset structure. Their analysis shows poor transfer of many post-hoc methods, with deep feature--based approaches performing more robustly \cite{gungor2025ts}. Beyond this, most \ac{ood} methods are developed for vision or language and are sparsely evaluated on time series, with only a few dedicated approaches such as SRS \cite{belkhouja2023out} and Diversify \cite{lu2024diversify}. Overall, the literature lacks systematic studies of hyperspherical representations for time-series \ac{ood} detection, comprehensive cross-dataset evaluations, and principled auxiliary-data settings.

\section{Preliminaries} 
\label{sec:problem-definition}
 
\subsection{Problem Statement}
\Ac{ood} detection aims to identify test inputs for which model predictions are unreliable, namely samples outside the intended generalization scope of the model \cite{zhang2023openood}. It is commonly formulated as a binary decision problem in which a test sample $x$ is determined to originate either from the \ac{id} distribution $P_{\text{in}}$ or from an unknown \ac{ood} distribution $P_{\text{out}}$. In standard image-classification benchmarks, for example, CIFAR-10 may be used as $P_{\text{in}}$, while CIFAR-100 or TinyImageNet are used as near-\ac{ood} data, and Places365 or SVHN are used as far-\ac{ood} data.

Let $\mathcal{Y}_{\text{in}}$ denote the \ac{id} label space, and let $\mathcal{D}_{\text{test}}^{\text{ood}}$ be an \ac{ood} test set with label space $\mathcal{Y}_{\text{ood}}$ such that $\mathcal{Y}_{\text{ood}} \cap \mathcal{Y}_{\text{in}}=\emptyset$. For an \ac{id} input $x$, the classifier $f_{\theta}$ provides class-posterior estimates $p_{\theta}(y \mid x)$ for $y \in \mathcal{Y}_{\text{in}}$ and assigns the label
\[
  \hat{y}=\arg\max_{y \in \mathcal{Y}_{\text{in}}} p_{\theta}(y \mid x).
\]
Thus, the decision problem is dual: samples drawn from $P_{\text{in}}$ must be classified over $\mathcal{Y}_{\text{in}}$, whereas samples drawn from $P_{\text{out}}$ must be detected and rejected. The latter decision is represented by an \ac{ood} score $S(\cdot;f_{\theta}):\mathcal{X}\rightarrow\mathbb{R}$, which induces the threshold detector
\[
  G_{\lambda}(x;f_{\theta}) = \mathds{1}\{ S(x;f_{\theta}) \geq \lambda \},
\]
where larger scores indicate higher \ac{ood} likelihood; for similarity-based scores, such as cosine similarity, the negated similarity is used to preserve this convention. The threshold $\lambda$ is usually selected so that a fixed proportion of \ac{id} samples, e.g., $95\%$, are accepted as \ac{id}.

\begin{figure*}[!t]
    \centering
    \includegraphics[width=0.75\linewidth]{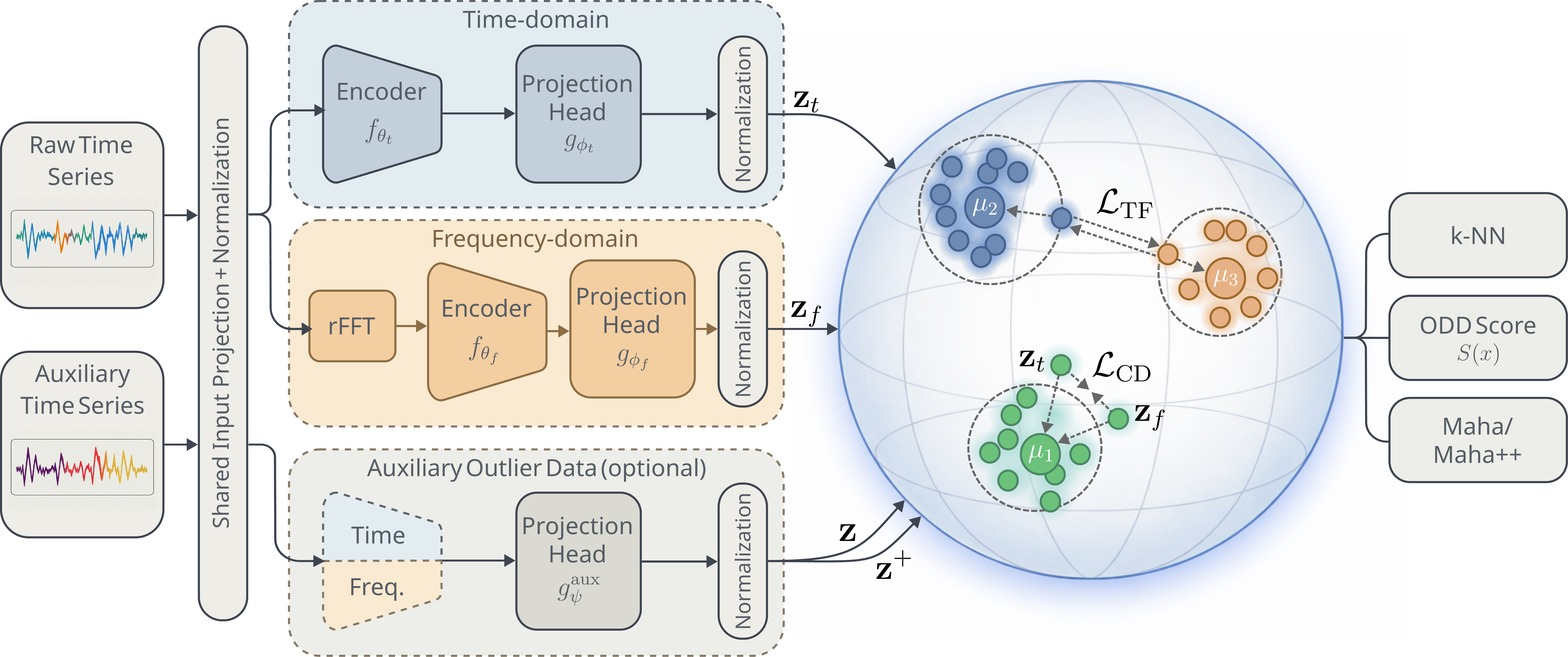}
    \caption{Architecture of the proposed HyperTF model. The input sequence is projected into a shared latent space and processed in parallel by time- and frequency-domain encoders. Domain-specific embeddings are mapped onto a shared hyperspherical classifier with common class prototypes.}
    \label{fig:framework}
\end{figure*}

\subsection{Hyperspherical Embeddings}
\label{subsec:hyperspherical-embeddings}

We adopt a hyperspherical representation framework in which normalized class-conditional features lie on the unit sphere and are modeled using directional probability distributions. Let $\mathbb{S}^{d-1} := \{ \mathbf{u} \in \mathbb{R}^{d} : \|\mathbf{u}\|_2 = 1 \}$ denote the unit $(d-1)$-sphere, and let $\mathbf{z} \in \mathbb{S}^{d-1}$ represent a normalized feature vector. For a classification problem with class set $\mathcal{C} := \{1,\dots,C\}$, each class $c \in \mathcal{C}$ is associated with a prototype direction $\boldsymbol{\mu}_c \in \mathbb{S}^{d-1}$.

Following prior work on hyperspherical feature modeling \cite{ming2023cider,burapacheepyour}, the conditional distribution of $\mathbf{z}$ given label $y=c$ is modeled by a \ac{vmf} density:
\begin{equation}
\label{eq:vmf-likelihood}
p(\mathbf{z} \mid y=c)
=
C_d(\kappa)\,
\exp\!\left(
\kappa\,\boldsymbol{\mu}_c^\top \mathbf{z}
\right),
\end{equation}
where $\kappa > 0$ controls angular dispersion and $C_d(\kappa)$ is the normalization constant. Under this model, classification depends only on the angular similarity between $\mathbf{z}$ and the class prototypes.

\section{Methodology}
\label{sec:methodology}

This section formalizes the proposed HyperTF framework (Figure~\ref{fig:framework}). The model employs a dual-stream architecture in the time and frequency domains, where augmented inputs are projected into a shared hyperspherical embedding space and supervised by common class prototypes. Training minimizes a composite objective that combines supervised hyperspherical classification, cross-domain feature alignment, and an optional auxiliary contrastive regularizer:
\[
\mathcal{L}
=
\mathcal{L}_{\mathrm{TF}}
+
\lambda_{\mathrm{CD}}\,\mathcal{L}_{\mathrm{CD}}
+
\lambda_{\mathrm{AX}}\,\mathcal{L}_{\mathrm{AX}},
\]
where $\mathcal{L}_{\mathrm{TF}}$, $\mathcal{L}_{\mathrm{CD}}$, and $\mathcal{L}_{\mathrm{AX}}$ are defined in Equations~\eqref{eq:loss_sup}, \eqref{eq:loss_consistency}, and~\eqref{eq:loss_aux}, and $\lambda_{\mathrm{CD}}$ and $\lambda_{\mathrm{AX}}$ weight the regularization terms. For simplicity, we set $\lambda_{\mathrm{CD}} = \lambda_{\mathrm{AX}} = 1$ in all experiments.

\subsection{Model Overview and Notation}
\label{subsec:model-overview}

Let $x \in \mathbb{R}^{S \times F}$ denote an input sequence, where $S$ and $F$ index the temporal length and channel dimensionality, respectively. During training, stochastic view sampling is performed using an augmentation distribution $\mathcal{A}$, producing paired samples $x^{(1)}, x^{(2)} \sim \mathcal{A}(x)$. Since the model is designed to be view-consistent, subsequent mappings are defined without explicitly indexing the views. An initial channel projection $\Pi : \mathbb{R}^{S \times F} \rightarrow \mathbb{R}^{S \times D_{\mathrm{proj}}}$ is applied independently at each time step, yielding a projected sequence $\hat{x} = \Pi(x)$; $\Pi$ is implemented as a linear embedding followed by normalization.

From the projected sequence $\hat{x}$, two complementary but aligned representations are constructed. The time-domain representation is defined directly as the projected signal, while the frequency-domain representation is obtained by applying a real-valued Fourier transform along the temporal dimension followed by a logarithmic magnitude scaling, yielding
\[
v_t \coloneqq \hat{x}, \qquad
v_f \coloneqq \log \bigl| \mathrm{rFFT}(\hat{x}) \bigr|.
\]
These representations are processed by parallel, domain-specific encoders,
\[
\mathbf{h}_t = f_{\theta_t}(v_t), \qquad
\mathbf{h}_f = f_{\theta_f}(v_f),
\]
which share the same architectural design but use separate parameter sets. Each encoder aggregates temporal information through pooling operations (mean+max concatenation), resulting in fixed-dimensional representations $\mathbf{h}_t, \mathbf{h}_f \in \mathbb{R}^d$. To apply shared prototype-based supervision without imposing cross-domain invariance, the encoder outputs are mapped by domain-specific supervised projection heads,
\[
\tilde{\mathbf{z}}_t = g_{\phi_t}(\mathbf{h}_t), \qquad
\tilde{\mathbf{z}}_f = g_{\phi_f}(\mathbf{h}_f).
\]
The resulting embeddings reside in a common hyperspherical space while retaining domain-specific structure for downstream classification.

Each projected feature is $\ell_2$-normalized, producing unit-norm embeddings $\mathbf{z}_t, \mathbf{z}_f \in \mathbb{S}^{d-1}$. 
A shared set of learnable class prototypes $\{ \boldsymbol{\mu}_c \}_{c=1}^C \subset \mathbb{S}^{d-1}$ defines a common angular decision space across domains. 
At inference time, class scores are obtained by averaging time- and frequency-domain prototype similarities:
\[
\ell_c
\;\coloneqq\;
\tau^{-1}
\boldsymbol{\mu}_c^\top
\left(
\tfrac{1}{2}(\mathbf{z}_t + \mathbf{z}_f)
\right),
\]
where $\tau > 0$ is a temperature parameter. 
During training, $\mathbf{z}_t$ and $\mathbf{z}_f$ are supervised separately using the same prototypes, so that both branches remain discriminative while retaining domain-specific information.

\subsection{Shared Hyperspherical Supervision via von Mises--Fisher Modeling}
\label{subsec:sup-hyperspherical}

Building on the hyperspherical embedding framework introduced in Section~\ref{subsec:hyperspherical-embeddings}, this subsection derives the supervised classification objective used to train the proposed model.
Let $\mathbf{z} \in \mathbb{S}^{d-1}$ denote a unit-norm embedding produced by either processing branch, and let $\{ \boldsymbol{\mu}_c \}_{c=1}^C \subset \mathbb{S}^{d-1}$ denote the shared set of learnable class prototypes.
Assuming the class-conditional \ac{vmf} model defined in Equation~\eqref{eq:vmf-likelihood}, the likelihood of $\mathbf{z}$ given label $y=c$ is
\[
p(\mathbf{z} \mid y = c)
= C_d(\kappa)\,\exp \bigl(\kappa\,\boldsymbol{\mu}_c^\top \mathbf{z}\bigr),
\]
where $\kappa > 0$ denotes the concentration parameter and $C_d(\kappa)$ is the corresponding normalization constant.

Under the assumption of a uniform class prior, application of Bayes’ rule yields a posterior distribution of the form
\[
\mathbb{P}(y=c \mid \mathbf{z})
= \frac{\exp \bigl(\kappa\,\boldsymbol{\mu}_c^\top \mathbf{z}\bigr)}{\sum_{c'} \exp \bigl(\kappa\,\boldsymbol{\mu}_{c'}^\top \mathbf{z}\bigr)},
\]
which coincides with a cosine-softmax classifier. Introducing a temperature parameter $\tau \coloneqq \kappa^{-1}$ recovers the scaled inner-product formulation employed in the previous subsection. Maximizing the conditional log-likelihood under this model leads to the standard cross-entropy objective; for a labeled in-distribution sample $(x,y)$ with embedding $\mathbf{z}$, the resulting loss is
\[
\mathcal{L}_{\mathrm{CE}}(\mathbf{z}, y)
= - \tau^{-1}\boldsymbol{\mu}_y^\top \mathbf{z}
+ \log \sum_{c'} \exp \left(\tau^{-1}\boldsymbol{\mu}_{c'}^\top \mathbf{z}\right).
\]

\begin{table*}[t]
\centering
\scriptsize
    \begin{tabularx}{\textwidth}{%
            >{\raggedright\arraybackslash}p{3.1cm}  
            >{\centering\arraybackslash}p{1.5cm}   
            *{6}{>{\centering\arraybackslash}p{1.5cm}}  
            >{\centering\arraybackslash}p{0.5cm}   
        }
        \toprule
        \multirow{2.4}{*}{\textbf{Method}} & \multirow{2.4}{*}{\textbf{Near-OOD}} & \multicolumn{6}{c}{\textbf{Far-OOD}} & \multirow{2.4}{*}{\textbf{F1}} \\

        \cmidrule(lr){3-8}
        & & \textbf{HAR} & \textbf{AUDIO} & \textbf{MOTION} & \textbf{ECG} & \textbf{SPECTRO} & \textbf{OTHER} & \\ 
        \midrule
        \multicolumn{9}{c}{\textbf{UCR (Univariate)}} \\
        \midrule
        MSP 
        & 54.45 / 66.99 & 55.69 / 60.69 & 43.38 / 79.75 & 42.22 / 78.48 & 70.01 / 45.49 & 38.82 / 78.51 & 49.87 / 67.16 & 78.99 \\
        Mahalanobis 
        & 91.90 / 15.83 & 97.03 / 6.04 & 74.73 / 40.47 & 84.85 / 23.35 & 97.03 / 4.48 & 99.56 / 0.92 & 95.81 / 6.11 & 78.99 \\
        Center Loss 
        & 92.73 / 14.43 & 98.58 / 3.36 & 84.21 / 31.14 & 85.45 / 22.89 & 98.43 / 3.11 & 99.90 / 0.32 & 96.89 / 4.59 & 78.81\\
        SimCLR+CE 
        & 91.44 / 18.37 & 96.31 / 9.23 & 66.03 / 51.27 & 88.86 / 23.92 & 98.71 / 3.97 & 99.03 / 2.76 & 94.82 / 8.86 & 72.16\\
        KNN+ 
        & 93.14 / 14.45 & 96.50 / 8.88 & 91.49 / 24.16 & 93.82 / 14.46 & 99.69 / 0.89 & 99.70 / 0.49 & 96.12 / 5.20 & 67.49 \\
        SSD+ (w/ Maha++) 
        & 93.83 / 12.93 & 97.85 / 5.57 & 92.99 / 17.24 & 92.01 / 15.10 & \textbf{99.76 / 0.64} & 99.64 / 0.49 & 97.08 / 3.89 & 67.49 \\
        CIDER
        & 92.78 / 15.38 & 96.38 / 9.01 & 89.20 / 27.53 & 93.15 / 16.15 & 99.41 / 1.61 & 99.61 / 1.04 & 96.53 / 5.65 & 78.88 \\ 
        \midrule
        HyperTF 
        & 93.94 / 12.91 & 99.13 / 2.42 & 94.63 / 13.61 & \textbf{93.91 / 13.53} & 99.48 / 1.34 & 99.95 / 0.19 & 99.40 / 1.41 & \textbf{82.53} \\ 
        + Auxiliary Outlier Data        & \textbf{94.11 / 12.56} & \textbf{99.31 / 1.92} & \textbf{95.24 / 12.43} & 93.19 / 14.81 & 99.58 / 1.16 & \textbf{99.97 / 0.12} & \textbf{99.45 / 1.13} & 79.14 \\
        \midrule

        \multicolumn{9}{c}{\textbf{UEA (Multivariate)}} \\
        \midrule
        MSP 
        & 55.19 / 62.54 & 49.22 / 65.79 & 14.82 / 96.95 & 72.58 / 41.89 & 32.18 / 81.59 & 38.96 / 61.56 & 61.02 / 51.84 & 70.64 \\
        Mahalanobis 
        & 87.56 / 18.21 & 88.53 / 15.75 & \textbf{99.87 / 0.35} & 98.35 / 4.20 & 91.20 / 12.06 & 92.32 / 7.78 & 82.77 / 19.63 & 70.64 \\
        Center Loss 
        &  89.17 / 15.12 & 90.02 / 13.36 & 99.83 / 0.37 & 99.25 / 1.71 & 95.06 / 6.90 & 96.17 / 3.92 & 84.73 / 18.33 & 70.00 \\
        SimCLR+CE 
        & 90.35 / 18.57 & 94.08 / 11.09 & 92.18 / 14.95 & 99.50 / 1.39 & 94.66 / 11.87 & 89.85 / 10.92 & 90.87 / 12.60 & 76.14 \\
        KNN+ 
        & 91.62 / 12.31 & 92.44 / 10.30 & 97.96 / 5.06 & 99.71 / 0.62 & 96.52 / 7.30 & 54.74 / 45.85 & 91.07 / 9.81 & 70.16 \\
        SSD+ (w/ Maha++) 
        & 93.44 / 10.03 & 93.91 / 9.02 & 99.39 / 1.64 & 99.87 / 0.38 & \textbf{96.85 / 7.22} & 92.20 / 7.86 & 91.16 / 9.60 & 70.16 \\
        CIDER
        & 92.74 / 13.81 & 95.55 / 7.49 & 93.63 / 11.45 & 99.55 / 0.78 & 92.23 / 12.98 & 91.44 / 9.23 & 90.90 / 10.75 & 77.75 \\
        \midrule
        HyperTF 
        & 97.76 / 6.05 & 96.61 / 5.56 & 98.65 / 2.72 & 99.94 / 0.31 & 95.24 / 10.91 & 99.77 / 0.26 & \textbf{99.55 / 1.32} & \textbf{81.74} \\ 
        + Auxiliary Outlier Data        & \textbf{98.23 / 4.94} & \textbf{97.12 / 4.54} & 98.47 / 2.66 & \textbf{99.97 / 0.16} & 95.52 / 10.67 & \textbf{99.88 / 0.16} & 96.16 / 5.55 & 80.73 \\
        \bottomrule
    \end{tabularx}
    \caption{\Ac{ood} detection performance on datasets from the UCR and UEA archives. Results are reported separately for \textit{near}-\ac{ood} and \textit{far}-\ac{ood} settings, with the latter further decomposed into the five largest modalities; remaining modalities are grouped under \textit{OTHER}. Metrics are \ac{auroc}$\uparrow$/\ac{fpr95}$\downarrow$ (\%), with the final column reporting the \ac{id} F1-score$\uparrow$ (\%).}
    \label{tab:main_exp}
\end{table*}

Supervision is applied independently to embeddings produced by the time- and frequency-domain encoders, yielding the combined supervised objective
\begin{equation}\label{eq:loss_sup}
\mathcal{L}_{\mathrm{TF}}
= \mathbb{E}_{(x,y)} \left[
\mathcal{L}_{\mathrm{CE}}(\mathbf{z}_t, y)
+ \mathcal{L}_{\mathrm{CE}}(\mathbf{z}_f, y)
\right].
\end{equation}
This objective promotes intra-class compactness across both domains, while implicitly encouraging inter-class separation through competition on the hypersphere. Consequently, discriminative structure is induced within each domain-specific embedding space.

\subsection{Cross-Domain Feature Consistency}
\label{subsec:cross-domain-consistency}

While the supervised objective in Equation~\eqref{eq:loss_sup} constrains class-conditional structure within each domain, it leaves the relative alignment of domain-specific representations underdetermined. To regularize this degree of freedom, a cross-domain consistency term is introduced at the level of the normalized embeddings. Let $\mathbf{z}_t, \mathbf{z}_f \in \mathbb{S}^{d-1}$ denote the $\ell_2$-normalized outputs of the supervised projection heads. The consistency objective is defined as
\begin{equation}\label{eq:loss_consistency}
    \mathcal{L}_{\mathrm{CD}}
    \;\coloneqq\;
    \mathbb{E}_{x}
    \bigl[
    1 - \mathbf{z}_t^\top \mathbf{z}_f
    \bigr],
\end{equation}
where the inner product corresponds to cosine similarity on the unit hypersphere and induces an angular alignment objective between the two domains.

\subsection{Auxiliary Contrastive Regularization}
\label{subsec:auxiliary-contrastive}

Training may optionally leverage unlabeled or weakly specified auxiliary time-series from $P_{\mathrm{aux}}$, whose support need not coincide with $P_{\mathrm{in}}$. For $x \sim P_{\mathrm{aux}}$, the augmentation operator $\mathcal{A}$ produces two correlated views $x$ and $x^{+}$, which are processed by the shared input projection and domain-specific encoders. A shared auxiliary projection head $g^{\mathrm{aux}}_{\psi}$ then maps the branch-specific representations of the two augmented views to normalized auxiliary embeddings $\mathbf{z}, \mathbf{z}^{+} \in \mathbb{S}^{d-1}$, which form an anchor-positive pair.

The auxiliary embeddings optimize a contrastive objective in which each anchor is contrasted against auxiliary and \ac{id} negatives. This mitigates collapse via instance discrimination while encouraging auxiliary--\ac{id} separation. Since gradients propagate through the auxiliary head into the encoders, the loss regularizes the representation geometry used for downstream \ac{ood} detection. The auxiliary contrastive loss is defined as
\begin{equation}
\label{eq:loss_aux}
\resizebox{.91\linewidth}{!}{$
\displaystyle
\mathcal{L}_{\mathrm{AX}}
=
\mathbb{E}_{\substack{x \sim P_{\mathrm{aux}}\\ x, x^{+} \sim \mathcal{A}(x)}}
\left[
-\log
\frac{
\exp \bigl( \tau^{-1}\mathbf{z}^{\top}\mathbf{z}^{+} \bigr)
}{
\exp \bigl( \tau^{-1}\mathbf{z}^{\top}\mathbf{z}^{+} \bigr)
+ Z(\mathbf{z}; M_{\mathrm{aux}}, M_{\mathrm{in}})
}
\right]
$}
\end{equation}
where $\tau > 0$. The normalization term aggregates similarity scores with respect to negative samples drawn from both auxiliary and in-distribution sources. In particular, let $\{\tilde{x}_m\}_{m=1}^{M_{\mathrm{aux}}}$ be i.i.d.\ draws from $P_{\mathrm{aux}}$ with augmented views $\tilde{x}'_m \sim \mathcal{A}(\tilde{x}_m)$, and let $\{x^{-}_j\}_{j=1}^{M_{\mathrm{in}}}$ be i.i.d.\ draws from $P_{\mathrm{in}}$, producing embeddings $\tilde{\mathbf{z}}'_m$ and $\mathbf{z}^{-}_j$, respectively. Then
\[
\resizebox{.95\linewidth}{!}{$
\displaystyle
Z(\mathbf{z}; M_{\mathrm{aux}}, M_{\mathrm{in}})
=
\sum_{m=1}^{M_{\mathrm{aux}}}
\exp \bigl( \tau^{-1}\mathbf{z}^\top \tilde{\mathbf{z}}'_m \bigr)
+
\sum_{j=1}^{M_{\mathrm{in}}}
\exp \bigl( \tau^{-1}\mathbf{z}^\top \mathbf{z}^{-}_j \bigr).
$}
\]
where $M_{\mathrm{aux}} + M_{\mathrm{in}} = M - 1$, and the \ac{id} negatives ${\mathbf{z}^{-}}$ are treated as fixed reference embeddings in $\mathcal{L}_{\mathrm{AX}}$.

\section{Experiments}\label{sec:experiments}
\subsection{Experimental Setup}
\paragraph{Datasets and training.} 
Experiments are conducted on the full UCR Time Series Classification Archive and the UEA Multivariate Time Series Classification Archive. Each dataset is treated in turn as the \ac{id}, with all remaining datasets serving as \ac{ood}. This results in a large-scale evaluation protocol covering 158 datasets in total and a broad range of sensing modalities; the benchmark protocol is described in Appendix~B. Unless stated otherwise, models are trained independently for each \ac{id} dataset, and results are averaged. All methods employ an \textsc{InceptionTime}~\cite{ismail2020inceptiontime} backbone followed by a projection head that maps representations into a $128$-dimensional $\ell_2$-normalized embedding space. Optimization is performed using \ac{sgd} with momentum $0.9$ and weight decay $3 \times 10^{-3}$ for $100$ epochs, using a batch size of $8$~\cite{yue2022ts2vec}. A cosine learning rate schedule is applied throughout training, and the temperature parameter is fixed to $\tau = 0.025$. Further details regarding architectural variants, data augmentations, and computational resources are deferred to Appendix~A. Dataset-level statistics and modality annotations are provided in Appendix~D.

\paragraph{\Ac{ood} detection scores.} 
\Ac{ood} detection is performed via distance-based scoring functions defined over the concatenated temporal and frequency learned embeddings. We consider three scoring strategies. First, we employ a \ac{knn} score with $k=1$, based on cosine similarity to the closest \ac{id} training embedding. Second, we use the classical Mahalanobis score~\cite{lee2018simple_maha}, computed using class-conditional mean and covariance estimates of \ac{id} embeddings. Finally, we include \ac{maha}~\cite{mueller2025mahalanobis++}, which applies $\ell_2$ normalization to the feature representations prior to covariance estimation, thereby mitigating violations of the Gaussian assumption caused by feature norm variability and yielding a more stable distance-based score.

\paragraph{Evaluation metrics.}  
Performance is assessed using three standard metrics for \ac{ood} detection and \ac{id} classification. Specifically, we report: (i) \ac{fpr95}, which measures the fraction of \ac{ood} samples incorrectly classified as \ac{id} when \ac{id} recall is fixed at $95\%$; (ii) \ac{auroc}, which captures threshold-independent separation between \ac{id} and \ac{ood} samples; and (iii) the \textsc{F1}-score for \ac{id} classification, which reflects classification performance. 

\subsection{Main Results}\label{sec:main_results}
We evaluate a broad spectrum of competitive baselines. Post-hoc approaches such as \ac{msp}~\cite{hendrycks2017msp} and Mahalanobis~\cite{lee2018simple_maha} rely on standard softmax \ac{ce} training, while Center Loss~\cite{yandong2016centerloss} explicitly promotes intra-class compactness in the embedding space. For contrastive learning--based methods, we include SimCLR~\cite{chen2020simple} combined with \ac{ce} and \ac{knn} scoring, as well as SSD+~\cite{sehwag2021ssd} and KNN+~\cite{sun2022knn}, both of which employ the SupCon objective. We further include CIDER~\cite{ming2023cider}, which optimizes hyperspherical embeddings via EMA-updated prototypes and explicitly encourages inter-prototype separation through a separation loss. All methods are evaluated using the same backbone architecture and embedding dimensionality.

Table~\ref{tab:main_exp} reports \ac{ood} detection performance on the UCR and UEA benchmarks, spanning both \textit{near}-\ac{ood} and a diverse set of \textit{far}-\ac{ood} scenarios across multiple modalities. HyperTF with auxiliary outlier data achieves the strongest near-\ac{ood} detection results among all evaluated approaches on both UCR and UEA, and HyperTF variants obtain the best results in most far-\ac{ood} groups. The near-\ac{ood} scenarios consist of datasets from the same modality as the \ac{id} data, often sharing sensing, preprocessing, and temporal characteristics, and therefore primarily reflect semantic rather than low-level distribution shifts; strong performance in this regime indicates sensitivity to subtle within-modality deviations. For far-\ac{ood} detection, HyperTF provides more consistent performance across heterogeneous modalities; although SSD+ remains competitive in \ac{ecg} and Mahalanobis is strongest on UEA AUDIO, our approach improves most groups overall. In addition, HyperTF substantially enhances \ac{id} classification accuracy, achieving an F1-score of $82.53$ on UCR compared to $78.88$ for CIDER, and $81.74$ on UEA versus $77.75$ for CIDER, indicating that improved \ac{ood} robustness does not come at the expense of \ac{id} performance.

\begin{table}[t]
    \centering
    \scriptsize
    \begin{tabularx}{\linewidth}{>{\hsize=1.5cm}X >{\hsize=0.7cm}C *{2}{C} *{2}{C}}
    \toprule
    \multirow{2.4}{*}{\textbf{Method}} 
    & \multirow{2.4}{*}{\textbf{Scoring}} 
    & \multicolumn{2}{c}{\textbf{UCR}} 
    & \multicolumn{2}{c}{\textbf{UEA}} \\
    \cmidrule(lr){3-4} \cmidrule(lr){5-6}
    & & \textbf{Near} & \textbf{Far} & \textbf{Near} & \textbf{Far} \\
    \midrule
    OE & MSP & 72.59 & 69.23 & 76.06 & 69.25 \\
    \midrule
    \multirow{2.4}{*}{Binary} 
    & $k$-NN & 16.17 & 5.67 & 16.90 & 13.01 \\
    & Maha++   & 13.28 & 3.07 & 13.27 & 7.36 \\
    \multirow{2.4}{*}{HyperTF} 
    & $k$-NN & 15.18 & 4.46 & 7.62 & 7.40 \\
    & Maha++   & 12.56 & \textbf{1.90} & 4.94 & 4.21 \\
    \multirow{2.4}{*}{HyperTF+MixOE} 
    & $k$-NN & 15.01 & 4.34 & 6.98 & 5.30 \\
    & Maha++   & \textbf{12.54} & 1.92 & \textbf{4.88} & \textbf{3.03} \\
    \bottomrule
    \end{tabularx}
    \caption{{ \Ac{ood} detection performance (\ac{fpr95} $\downarrow$) under different strategies for incorporating auxiliary outlier data during training, evaluated on Near- and Far-\ac{ood} settings for UCR and UEA.}}
    \label{tab:aux-study}
\end{table} 

\begin{table}[t]
    \centering
    \scriptsize
    \begin{tabularx}{\linewidth}{>{\hsize=1.5cm}X >{\hsize=0.7cm}C *{2}{C} *{2}{C}}
    \toprule
    \multirow{2.4}{*}{\textbf{Method}} 
    & \multirow{2.4}{*}{\textbf{Scoring}} 
    & \multicolumn{2}{c}{\textbf{UCR}} 
    & \multicolumn{2}{c}{\textbf{UEA}} \\
    \cmidrule(lr){3-4} \cmidrule(lr){5-6}
    & & \textbf{Near} & \textbf{Far} & \textbf{Near} & \textbf{Far} \\
    \midrule
    KNN+ & $k$-NN  & 14.45 & 5.71 & 12.31 & 8.97 \\
    SSD+ & Maha++  & 12.93 & 4.30 & 10.03 & 6.66 \\
    \midrule
    \multirow{2.4}{*}{HyperTF} 
                & $k$-NN  & 16.76 & 6.04 & 8.60 & 5.99 \\
                & Maha++  & \textbf{12.91} & \textbf{2.12} & \textbf{6.05} & \textbf{3.30} \\
    \bottomrule
    \end{tabularx}
    \caption{{ \Ac{ood} detection performance (\ac{fpr95} $\downarrow$) comparing HyperTF with supervised contrastive baselines under their standard scoring rules on Near- and Far-\ac{ood} settings for UCR and UEA.}}
    \label{tab:score-study}
\end{table}

\begin{table*}[t]
    \centering
    \scriptsize
    \scriptsize
    \begin{tabularx}{\textwidth}{@{} C *{8}{C} @{}}
    \toprule
    \multirow{2.4}{*}{\textbf{Set}} & \multicolumn{8}{c}{\textbf{OOD Scores}} \\
    \cmidrule(lr){2-9}
     & Prototype & Energy & GEN & Maha & Maha++ & MaxLogit & MSP & kNN  \\
    \midrule 
    \textbf{Near} & 94.11 / 12.57 & 54.34 / 72.38 & 49.56 / 80.45 & 96.43 / 7.61 & \textbf{98.23 / 4.94} & 47.39 / 84.02 & 46.85 / 84.57 & 96.83 / 7.62 \\
    \midrule
    
    \textbf{Far} & 96.56 / 6.50 & 53.94 / 66.84 & 48.78 / 75.67 & 96.65 / 5.12 & \textbf{97.42 / 4.21} & 49.95 / 75.83 & 49.41 / 76.35 & 94.97 / 7.40  \\
    \bottomrule
    \end{tabularx}
    \caption{{ Comparison of post-hoc scoring methods while training with our proposed framework with the presence of auxiliary data.}}
    \label{tab:ood_aux_contrast_single}
\end{table*}
\begin{table*}[t]
    \centering
    \scriptsize
    \renewcommand{\arraystretch}{1.05}
    \begin{tabularx}{\textwidth}{@{}
    >{\hsize=3.5cm}X  
    >{\hsize=1.5cm}C  
    *{2}{CC}  
    @{}}
    \toprule
    \multirow{2.4}{*}{\textbf{Variant}} & \multirow{2.4}{*}{\textbf{Included}} & \multicolumn{2}{c}{\textbf{Near-OOD}} & \multicolumn{2}{c}{\textbf{Far-OOD}} \\
    \cmidrule(lr){3-4} \cmidrule(lr){5-6}
     &  & \textbf{kNN} & \textbf{Maha++} & \textbf{kNN} & \textbf{Maha++} \\
    \midrule 
    Euclidean CE & \xmark & 85.80 / 20.71 & 93.49 / 11.22 & 87.97 / 14.73 & 95.33 / 6.56  \\
    Hyperspherical CE & \cmark & 92.86 / 11.64 & 94.58 / 9.05 & 93.98 / 8.35 & 95.78 / 5.90  \\
    \quad + Two Views (Crop \& Resize) & \cmark & 92.91 / 11.87 & 94.74 / 8.83 & 93.75 / 8.61 & 95.78 / 6.07  \\
    \quad + Input Projection & \cmark & 95.69 / 10.15 & 97.63 / 5.87 & 95.60 / 7.21 & 98.17 / 3.57  \\
    \quad\quad + Masking Projected Inputs & \xmark & 94.33 / 11.79 & 97.53 / 6.08 & 95.44 / 7.33 & 97.97 / 4.00  \\
    \quad + Compactness Loss (C) & \xmark & 95.51 / 10.80 & 97.55 / 5.92 & 96.12 / 6.70 & 98.34 / 3.25  \\
    \quad + Separation Loss (S) & \xmark & 93.78 / 12.46 & 97.28 / 6.54 & 95.71 / 7.16 & 98.24 / 3.43  \\
    \midrule
    \quad + FFT and TF Loss & \cmark & 95.89 / 8.95 & 97.47 / 6.94 & 96.55 / 5.25 & 97.78 / 3.35  \\
    \quad\quad + FFT Consistency & \cmark & 95.92 / 7.94 & 97.47 / 6.19 & 96.60 / 4.87 & 98.03 / 2.96   \\
    \bottomrule
    \end{tabularx}
    \caption{{ Step-wise ablation of our approach on the full UEA set. We report \ac{auroc}$\uparrow$/\ac{fpr95}$\downarrow$ for \ac{ood} detection with \ac{knn} and Mahalanobis scores.}}
    \label{tab:ablation-build-up}
\end{table*}

\begin{figure*}[!t]
    \centering
    \begin{tabular}{ccc}
    \includegraphics[width=0.25\textwidth]{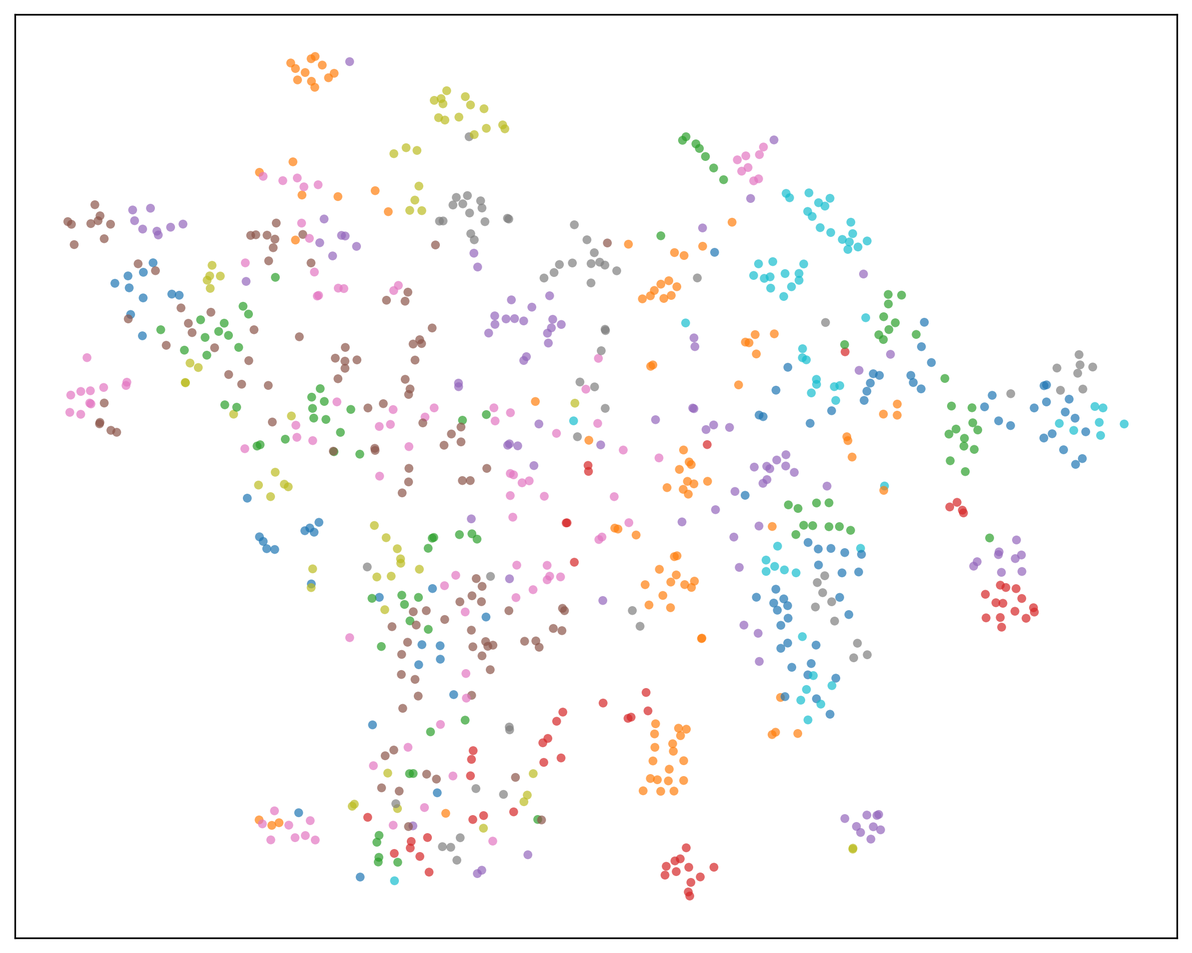} &
    \includegraphics[width=0.25\textwidth]{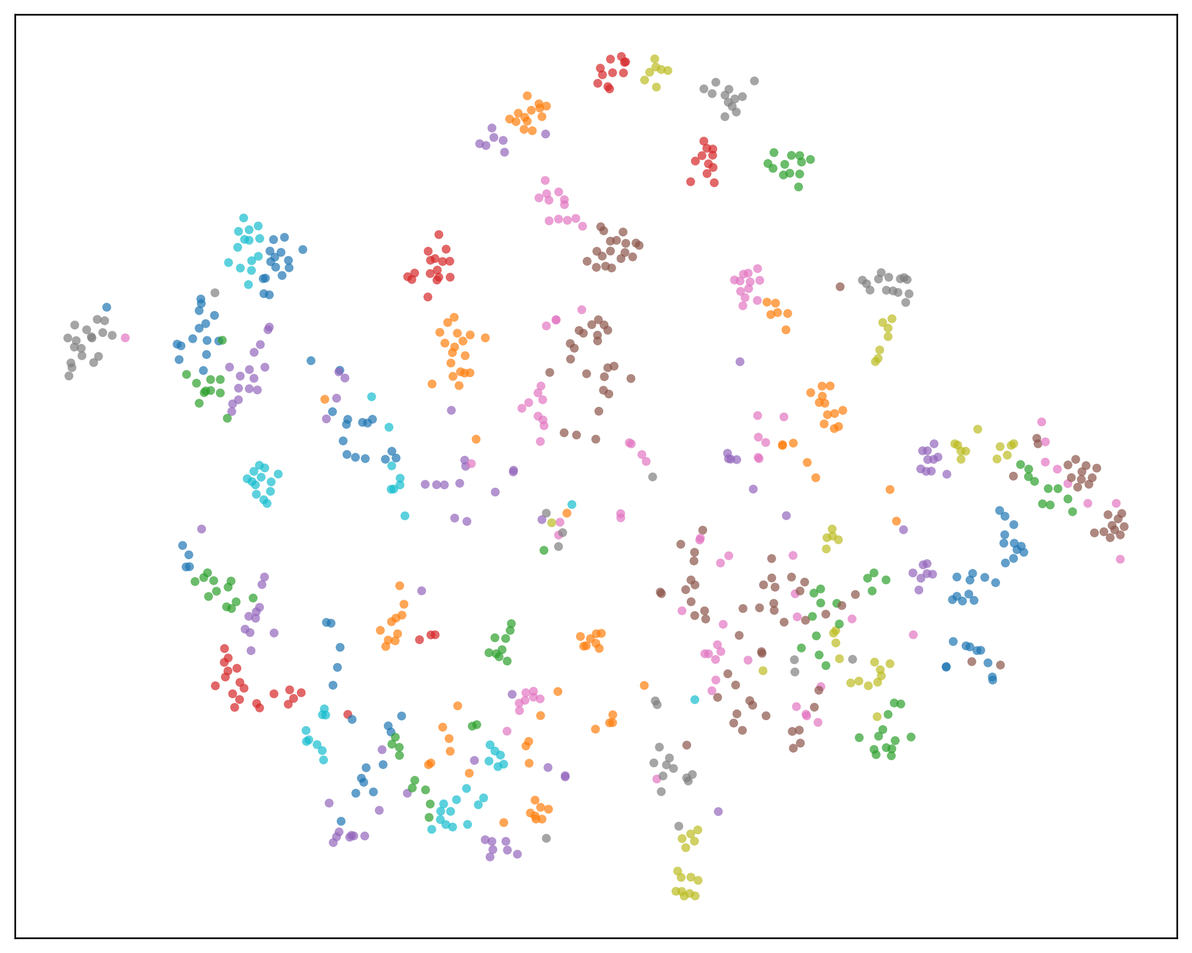} &
    \includegraphics[width=0.25\textwidth]{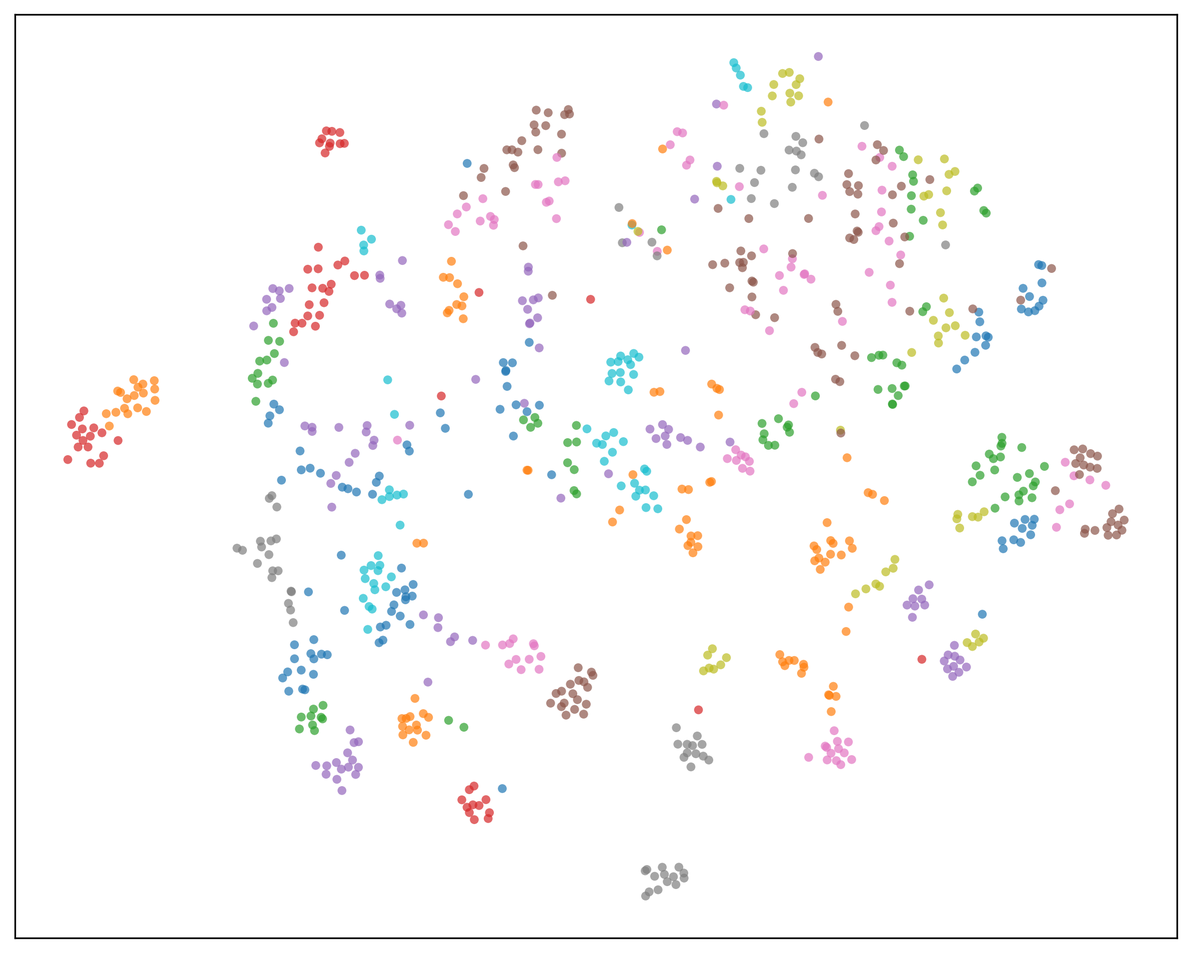}
    \end{tabular}
    \caption{t-SNE plots for the UEA Handwriting dataset: \ac{ce} (left), HyperTF (center), HyperTF with auxiliary outlier data (right).}
    \label{fig:tsne_three}
\end{figure*}

\paragraph{Auxiliary outlier data.}
We analyze the effect of incorporating auxiliary outlier data during training. Auxiliary datasets are selected to complement the evaluation setting: near-\ac{ood} performance is evaluated using models trained with far-\ac{ood} auxiliary data, and vice versa, ensuring that the model does not encounter the same type of distribution shift during both training and testing. Since far-\ac{ood} spans multiple modalities, its auxiliary pool is substantially larger than that used for near-\ac{ood}; as a result, far-\ac{ood} evaluation relies on a smaller and less diverse near-\ac{ood} auxiliary set. This setup reflects realistic scenarios in which available outlier data are limited and only weakly representative, motivating the additional evaluation of MixOE~\cite{zhang2023mixture} in combination with HyperTF to improve auxiliary data diversity during training.

Table~\ref{tab:aux-study} compares several ways of using auxiliary data: (i) a binary in-vs-outlier classifier; (ii) standard \ac{oe}~\cite{hendrycks2019oe}; and (iii) the proposed auxiliary contrastive regularizer with and without MixOE. The proposed contrastive formulation improves over the binary baseline across both UCR and UEA under matched scoring rules. On the UEA multivariate benchmark under near-\ac{ood} evaluation, the \ac{fpr95} decreases from $6.05$ without auxiliary data (Table~\ref{tab:main_exp}) to $4.94$ with auxiliary outliers, and further to $4.88$ with MixOE. The benefit of MixOE is most pronounced for UEA far-\ac{ood} evaluation, where \ac{fpr95} decreases from $4.21$ to $3.03$, indicating that increasing auxiliary data diversity is particularly beneficial when available outliers provide limited coverage.

\paragraph{Scoring functions.}
We next examine the interplay between our proposed training objective and post-hoc scoring functions. Table~\ref{tab:score-study} compares HyperTF with representative supervised contrastive baselines under their standard scoring rules, namely \ac{knn} for KNN+ and \ac{maha} for SSD+. Although SSD+ was originally evaluated with the Mahalanobis score, we adopt \ac{maha} for SSD+ in this study to ensure a fair comparison with HyperTF, and we find that this choice consistently improves SSD+ performance. Under the same backbone architecture and embedding dimensionality, HyperTF consistently attains lower \ac{fpr95} on both UCR and UEA. Notably, when evaluated with the same \ac{knn} scoring used by KNN+, HyperTF delivers markedly better performance on UEA, indicating that the observed gains are not driven solely by the choice of scoring function. Similarly, under \ac{maha} scoring, HyperTF matches or exceeds SSD+ on UCR and achieves clear improvements on UEA, particularly in far-\ac{ood} scenarios. Together, these results suggest that the proposed training objective induces representations with improved \ac{ood} separability, which translates into stronger detection performance under standard scoring schemes.

We further evaluate HyperTF with auxiliary data using standard post-hoc scoring methods (Section~\ref{sec:related_work}), including confidence-, distance-, and prototype-based approaches. Prototype-based scoring computes class means from \ac{id} embeddings and evaluates test inputs via cosine similarity to the nearest prototype. As shown in Table~\ref{tab:ood_aux_contrast_single}, distance-based methods, particularly \ac{maha}, perform best across both near- and far-\ac{ood} regimes.

\subsection{Ablation Studies}\label{sec:ablation_studies}

We study the contribution of individual components of HyperTF by progressively enabling each modeling choice and loss term.
Table~\ref{tab:ablation-build-up} presents a step-wise ablation that progressively enables the components of HyperTF, starting from a standard Euclidean softmax \ac{ce} baseline in the time domain. Replacing Euclidean logits with hyperspherical \ac{ce} based on cosine similarity to shared class prototypes yields a clear improvement in \ac{ood} detection. Subsequent additions of multi-view temporal crops and input projection further improve performance, with consistent gains observed in near-\ac{ood} detection using \ac{knn}. As a non-parametric metric, \ac{knn} provides a reliable indicator of representational quality, and its monotonic improvement across ablation steps reflects the progressive refinement of the learned embedding space. Auxiliary geometric regularizers have limited impact, with a compactness loss $(1 - \boldsymbol{\mu}_c^\top \mathbf{z})$ yielding only marginal gains, while separation loss~\cite{ming2023cider} provides no additional benefit. The lower block of Table~\ref{tab:ablation-build-up} extends the time-domain formulation by incorporating an \ac{fft}-based frequency-domain view, corresponding to the full $\mathcal{L}_{\mathrm{TF}}$ objective. Enforcing \ac{fft} consistency between time- and frequency-domain embeddings yields the strongest overall \ac{ood} detection performance.

Figure~\ref{fig:tsne_three} shows t-SNE visualizations of \ac{id} embeddings on the UEA Handwriting dataset. Moving from standard Euclidean \ac{ce} to HyperTF results in visibly greater intra-class compactness and inter-class separation. Incorporating auxiliary outlier data further sharpens class boundaries, indicating additional regularization of the embedding space.

\section{Conclusion}\label{sec:conclusion}  
This paper investigated \ac{ood} detection for time-series data through the lens of hyperspherical representation learning. We introduced HyperTF, a dual-view hyperspherical framework that unifies temporal and frequency-domain representation learning within a shared angular decision space, combining domain-specific encoders with shared prototype-based supervision and cross-domain consistency regularization. Across the complete UCR and UEA archives under a cross-dataset evaluation protocol, the proposed approach consistently improved \ac{ood} detection performance under standard distance-based scoring rules while also maintaining strong \ac{id} classification accuracy. These results suggest that jointly modeling complementary temporal and spectral structure within a shared geometric embedding space provides a robust foundation for time-series \ac{ood} detection under realistic distribution shifts. More broadly, our findings highlight the importance of evaluating time-series \ac{ood} methods beyond intra-dataset label splits and toward benchmark settings that better reflect deployment-time variation across sensing conditions, domains, and data sources.

\footnotesize
\bibliographystyle{named}
\bibliography{ijcai26}

\appendix
\onecolumn
\appendix

\section{Training Details}
\label{app:training_details}

We use \textsc{InceptionTime}~\cite{ismail2020inceptiontime} as the encoder network $f_\theta$, followed by a projection head $g_\phi$ that maps representations to a 128-dimensional hyperspherical embedding space. The projection head is implemented as a two-layer \ac{mlp} with batch normalization and ReLU activations (except the last one).  All models are trained by minimizing the total loss defined in Section~\ref{sec:methodology}. We use \ac{sgd} with momentum $0.9$ and weight decay $3 \times 10^{-3}$. The base learning rate is linearly warmed up over the first $10\%$ of training steps and subsequently annealed with a cosine schedule. Training is run for $100$ epochs per \ac{id} dataset, with a batch size of $8$, following~\cite{yue2022ts2vec}. For data augmentation, we apply a random crop in $[75\%, 99\%]$ and resize back to the original length using linear interpolation. When auxiliary outlier samples are incorporated during training, the running statistics of batch-normalization layers are updated exclusively using \ac{id} mini-batches. Auxiliary samples are processed without modifying these statistics, thereby preventing distributional contamination of the supervised feature space. In addition, within the auxiliary contrastive objective, \ac{id} embeddings used as negative references are treated as stop-gradient quantities; consequently, gradient propagation is restricted to the auxiliary outlier branch. All experiments are conducted on NVIDIA A100 GPUs. 

\section{Benchmark Protocol}
\label{app:benchmark_protocol}
 
Let $\mathcal{D} = \{D_1,\dots,D_N\}$ denote the collection of datasets drawn from the UCR/UEA time-series classification archives\footnote{\url{https://timeseriesclassification.com}}. 
Each dataset $D_i$ is treated as an empirical sample from an underlying data-generating distribution, with input space $\mathcal{X}_i \subset \mathbb{R}^{T_i \times d_i}$ and label space $\mathcal{Y}_i$, where $T_i$ and $d_i$ denote the sequence length and channel dimensionality (corresponding to $S$ and $F$ in the main text), respectively. 
Univariate UCR datasets correspond to $d_i = 1$, whereas multivariate UEA datasets satisfy $d_i > 1$. 

For each evaluation instance, one dataset $D_{\mathrm{ID}} \in \mathcal{D}$ is designated as the \ac{id} dataset. A model is trained exclusively on the canonical training partition of $D_{\mathrm{ID}}$, and its performance is subsequently evaluated on the corresponding test partition using the \ac{ood} scoring function $S:\mathcal{X}\rightarrow\mathbb{R}$. This procedure is repeated for each dataset in the archive, so that every $D_i \in \mathcal{D}$ serves as the \ac{id} dataset in turn. 
The remaining datasets constitute the candidate OOD pool,
$
    \mathcal{D}_{\mathrm{OOD}}
    := \mathcal{D} \setminus \{D_{\mathrm{ID}}\}.
$
To impose structured semantic relationships across datasets and to distinguish same-modality (near) from cross-modality (far) OOD conditions, we introduce a modality-mapping function
$
    \tau : \mathcal{D} \rightarrow \mathcal{T},
$
where $\mathcal{T}$ denotes a finite set of modality categories capturing coarse-grained semantic groupings such as \texttt{HAR}, \texttt{Motion}, \texttt{ECG}, \texttt{EOG}, \texttt{EEG}, \texttt{Audio}, \texttt{Spectro}, \texttt{Sensor}, and \texttt{Device}, among other physiologic or sensor-derived types. Full modality assignments for each UCR/UEA dataset appear in Tables~\ref{tab:ucr_datasets_1}, \ref{tab:ucr_datasets_2}, and \ref{tab:uea_datasets} (Appendix~\ref{app:dataset_statistics}). Certain modalities occur only once within the archives; to avoid the degenerate case in which no same-modality datasets are available for constructing near-\ac{ood} sets, singleton modalities are remapped to the closest semantic category using the rules in Tables~\ref{tab:ucr_modality_map} and~\ref{tab:uea_modality_map} (Appendix~\ref{app:modality_remap}).

\subsection{Auxiliary Outlier Data}
\label{subsec:aux-outlier-data}

Several contemporary approaches, most notably \ac{oe}~\cite{hendrycks2019oe}, incorporate an unlabeled auxiliary set during training to encourage low-confidence predictions on non-\ac{id} inputs. In the canonical \ac{oe} formulation, auxiliary samples may be drawn from any sufficiently broad distribution; disjointness from the test-time \ac{ood} distributions is not required, and partial overlap is explicitly permitted. 

In a benchmarking environment, however, this permissive construction introduces a methodological complication. Allowing auxiliary outliers to be drawn from the same datasets used for \ac{ood} evaluation would violate the separation-of-information principle, which is fundamental to fair assessment. Specifically, such overlap would enable a model to internalize dataset-specific signatures of the test-time \ac{ood} sources, thereby conflating genuine \ac{ood} generalization with the memorization of evaluation artifacts; consequently, estimates of \ac{auroc} and related metrics would be artificially inflated and non-comparable across studies. To preclude this leakage, our benchmark protocol enforces a strict partition between auxiliary and evaluation outliers by requiring that auxiliary samples be sourced from the \emph{complementary} \ac{ood} group relative to the group used at test time:
\[
  \mathcal{D}_{\mathrm{aux}}
  =
  \begin{cases}
    \mathcal{D}_{\mathrm{OOD}}^{\mathrm{far}}, &
      \text{if test-time evaluation uses }
      \mathcal{D}_{\mathrm{OOD}}^{\mathrm{near}}, \\[4pt]
    \mathcal{D}_{\mathrm{OOD}}^{\mathrm{near}}, &
      \text{if test-time evaluation uses }
      \mathcal{D}_{\mathrm{OOD}}^{\mathrm{far}}.
  \end{cases}
\]
Training therefore, proceeds on the augmented set
\[
    \mathcal{D}_{\mathrm{train}}
    = D_{\mathrm{ID}} \cup \mathcal{D}_{\mathrm{aux}},
\]
where all samples in $D_{\mathrm{ID}}$ retain their semantic labels in $\mathcal{Y}_{\mathrm{ID}} = \{1,\dots,C\}$. Auxiliary samples are unlabeled with respect to the \ac{id} task and are treated collectively as a single outlier pool, denoted $C+1$ for convenience. This notation is only conceptual and does not imply supervised training of an additional classifier class in the proposed method.

The complementary construction naturally induces asymmetric data regimes. When $\mathcal{D}_{\mathrm{aux}} = \mathcal{D}_{\mathrm{OOD}}^{\mathrm{near}}$, the auxiliary pool often contains fewer datasets and, consequently, fewer samples than its far-\ac{ood} counterpart. Although this configuration might appear to constrain auxiliary diversity, it is in fact beneficial for controlled experimentation. It permits systematic evaluation of \ac{oe}-style methods under both data-limited and data-rich auxiliary scenarios, thereby enabling analysis of (i) sensitivity to auxiliary pool size, (ii) dependence on regularization strength, and (iii) the effect of techniques such as MixOE that explicitly leverage auxiliary-sample abundance. 
\subsection{Shape Alignment for Cross-Dataset Evaluation}
\label{subsec:shape_alignment}

Models of the form $f_{\theta} : \mathbb{R}^{T \times d} \rightarrow \mathbb{R}^{k}$ require a fixed input shape, yet datasets in the UCR/UEA archives differ substantially in sequence length, sampling rate, and channel dimensionality. To enable cross-dataset \ac{ood} evaluation under a consistent architectural interface, all datasets are mapped deterministically into the canonical input space of the \ac{id} dataset $D_{\mathrm{ID}}$.

Let $D_{\mathrm{ID}}$ define the target shape $\mathbb{R}^{T_{\mathrm{ID}} \times d_{\mathrm{ID}}}$. For any OOD dataset $D_{\mathrm{OOD}}$ with original shape $\mathbb{R}^{T_{\mathrm{OOD}} \times d_{\mathrm{OOD}}}$, alignment proceeds through two operations: (i) \textit{temporal alignment}, each sequence is resampled to length $T_{\mathrm{ID}}$ via linear interpolation; (ii) \textit{channel alignment}, the resampled sequence is then mapped to $d_{\mathrm{ID}}$ channels. If $d_{\mathrm{OOD}} > d_{\mathrm{ID}}$, the dimensionality is reduced by truncating to the first $d_{\mathrm{ID}}$ channels, i.e., 
$
    \tilde{x}
    = \bigl( x_{t,j} \bigr)_{t = 1,\dots,T_{\mathrm{ID}},\; j = 1,\dots,d_{\mathrm{ID}}}.
$
If $d_{\mathrm{OOD}} < d_{\mathrm{ID}}$, channels are deterministically replicated until the dimensionality meets or exceeds $d_{\mathrm{ID}}$, after which the same index restriction yields exactly $d_{\mathrm{ID}}$ channels. The resulting representation satisfies
$
    \tilde{x} \in \mathbb{R}^{T_{\mathrm{ID}} \times d_{\mathrm{ID}}},
$
and is therefore directly compatible with the encoder $f_{\theta}$ and projection head $g_{\phi}$ introduced in Section~\ref{subsec:model-overview}.

\section{Dataset Splits and Evaluation Protocol}\label{app:dataset_splits_eval_proto}
All experiments rely on the canonical train/test splits provided by the UCR and UEA archives. For each dataset in the archive, we designate it as the \ac{id} source and train both our method and baselines exclusively on its training partition. After training, \ac{ood} scores are computed as follows: for the given $D_{\text{ID}}$, we evaluate against every other dataset $D_j \in \mathcal{D} \setminus \{D_{\text{ID}}\}$ separately, yielding a set of pairwise \ac{id}-vs-\ac{ood} results. For each \ac{id} dataset, we first average the \ac{ood} detection metrics across all $D_j$, while keeping near- and far-shift groups distinct. Finally, we aggregate the per-\ac{id} results by averaging across all datasets in the archive and report the mean values across multiple random trials. This ensures that every dataset is treated as a single \ac{id} and that results are statistically stable. The full list of \ac{id} datasets employed in our experiments, along with their types, which are critical for defining the near- and far-shift benchmarking splits, is provided in Tables~\ref{tab:ucr_datasets_1}, \ref{tab:ucr_datasets_2}, and~\ref{tab:uea_datasets}.

\section{Dataset Statistics}\label{app:dataset_statistics}
Our benchmark spans the complete set of datasets from the UCR and UEA archives. Since each dataset is treated in turn as an \ac{id} source for training, with all others serving as potential \ac{ood} targets, we provide full statistics covering train/test size, dimensionality, series length, and number of classes. The UCR collection is summarized in Tables~\ref{tab:ucr_datasets_1} and~\ref{tab:ucr_datasets_2}, and the UEA collection in Table~\ref{tab:uea_datasets}.

In addition to raw statistics, we also report a dataset \emph{type} label (e.g., \ac{har}, \ac{eeg}, Audio, etc.). This categorization is essential for our benchmark protocol, as it underlies the construction of near- vs. far-\ac{ood} splits: datasets belonging to the same type form natural candidates for near-\ac{ood} evaluation, while datasets from different types contribute to far-\ac{ood} scenarios.

\begin{table*}[ht]
    \centering\scriptsize
    \caption{UCR datasets used as \ac{id} in our experiments. The table reports dataset statistics along with a \textit{Type} label, which is used to define near- and far-\ac{ood} splits in our benchmark.}
    \label{tab:ucr_datasets_1}
    \begin{tabularx}{\columnwidth}{@{} >{\hsize=3cm}X CCCCCC @{}}
    \toprule
    \textbf{Dataset} & \textbf{TrainSize} & \textbf{TestSize} & \textbf{NumDimensions} & \textbf{SeriesLength} & \textbf{NumClasses} & \textbf{Type} \\
    \midrule
    ACSF1 & 100 & 100 & 1 & 1460 & 10 & DEVICE \\
    Adiac & 390 & 391 & 1 & 176 & 37 & IMAGE \\
    AllGestureWiimoteX & 300 & 700 & 1 & 500 & 10 & HAR \\
    AllGestureWiimoteY & 300 & 700 & 1 & 500 & 10 & HAR \\
    AllGestureWiimoteZ & 300 & 700 & 1 & 500 & 10 & HAR \\
    ArrowHead & 36 & 175 & 1 & 251 & 3 & IMAGE \\
    BME & 30 & 150 & 1 & 128 & 3 & SIMULATED \\
    Beef & 30 & 30 & 1 & 470 & 5 & SPECTRO \\
    BeetleFly & 20 & 20 & 1 & 512 & 2 & IMAGE \\
    BirdChicken & 20 & 20 & 1 & 512 & 2 & IMAGE \\
    CBF & 30 & 900 & 1 & 128 & 3 & SIMULATED \\
    Car & 60 & 60 & 1 & 577 & 4 & SENSOR \\
    Chinatown & 20 & 343 & 1 & 24 & 2 & TRAFFIC \\
    ChlorineConcentration & 467 & 3840 & 1 & 166 & 3 & SIMULATED \\
    CinCECGTorso & 40 & 1380 & 1 & 1639 & 4 & ECG \\
    Coffee & 28 & 28 & 1 & 286 & 2 & SPECTRO \\
    Computers & 250 & 250 & 1 & 720 & 2 & DEVICE \\
    CricketX & 390 & 390 & 1 & 300 & 12 & HAR \\
    CricketY & 390 & 390 & 1 & 300 & 12 & HAR \\
    CricketZ & 390 & 390 & 1 & 300 & 12 & HAR \\
    Crop & 7200 & 16800 & 1 & 46 & 24 & IMAGE \\
    DiatomSizeReduction & 16 & 306 & 1 & 345 & 4 & IMAGE \\
    DistalPhalanxOutlineAgeGroup & 400 & 139 & 1 & 80 & 3 & IMAGE \\
    DistalPhalanxOutlineCorrect & 600 & 276 & 1 & 80 & 2 & IMAGE \\
    DistalPhalanxTW & 400 & 139 & 1 & 80 & 6 & IMAGE \\
    DodgerLoopDay & 67 & 77 & 1 & 288 & 7 & SENSOR \\
    DodgerLoopGame & 17 & 127 & 1 & 288 & 2 & SENSOR \\
    DodgerLoopWeekend & 18 & 126 & 1 & 288 & 2 & SENSOR \\
    ECG200 & 100 & 100 & 1 & 96 & 2 & ECG \\
    ECG5000 & 500 & 4500 & 1 & 140 & 5 & ECG \\
    ECGFiveDays & 23 & 861 & 1 & 136 & 2 & ECG \\
    EOGHorizontalSignal & 362 & 362 & 1 & 1250 & 12 & EOG \\
    EOGVerticalSignal & 362 & 362 & 1 & 1250 & 12 & EOG \\
    Earthquakes & 322 & 139 & 1 & 512 & 2 & SENSOR \\
    ElectricDevices & 8926 & 7711 & 1 & 96 & 7 & DEVICE \\
    EthanolLevel & 504 & 500 & 1 & 1751 & 4 & SPECTRO \\
    FaceAll & 560 & 1690 & 1 & 131 & 14 & IMAGE \\
    FaceFour & 24 & 88 & 1 & 350 & 4 & IMAGE \\
    FacesUCR & 200 & 2050 & 1 & 131 & 14 & IMAGE \\
    FiftyWords & 450 & 455 & 1 & 270 & 50 & IMAGE \\
    Fish & 175 & 175 & 1 & 463 & 7 & IMAGE \\
    FordA & 3601 & 1320 & 1 & 500 & 2 & SENSOR \\
    FordB & 3636 & 810 & 1 & 500 & 2 & SENSOR \\
    FreezerRegularTrain & 150 & 2850 & 1 & 301 & 2 & DEVICE \\
    FreezerSmallTrain & 28 & 2850 & 1 & 301 & 2 & DEVICE \\
    Fungi & 18 & 186 & 1 & 201 & 18 & OTHER \\
    GestureMidAirD1 & 208 & 130 & 1 & 360 & 26 & HAR \\
    GestureMidAirD2 & 208 & 130 & 1 & 360 & 26 & HAR \\
    GestureMidAirD3 & 208 & 130 & 1 & 360 & 26 & HAR \\
    GesturePebbleZ1 & 132 & 172 & 1 & 455 & 6 & HAR \\
    GesturePebbleZ2 & 146 & 158 & 1 & 455 & 6 & HAR \\
    GunPoint & 50 & 150 & 1 & 150 & 2 & HAR \\
    GunPointAgeSpan & 135 & 316 & 1 & 150 & 2 & HAR \\
    GunPointMaleVersusFemale & 135 & 316 & 1 & 150 & 2 & HAR \\
    GunPointOldVersusYoung & 136 & 315 & 1 & 150 & 2 & HAR \\
    \bottomrule
    \end{tabularx}
\end{table*}

\begin{table*}[ht]
    \centering\scriptsize
    \caption{Continuation of Table~\ref{tab:ucr_datasets_1}.}
    \label{tab:ucr_datasets_2}
    \begin{tabularx}{\columnwidth}{@{} >{\hsize=3cm}X CCCCCC @{}}
    \toprule
    \textbf{Dataset} & \textbf{TrainSize} & \textbf{TestSize} & \textbf{NumDimensions} & \textbf{SeriesLength} & \textbf{NumClasses} & \textbf{Type} \\
    \midrule 
    Ham & 109 & 105 & 1 & 431 & 2 & SPECTRO \\
    HandOutlines & 1000 & 370 & 1 & 2709 & 2 & IMAGE \\
    Haptics & 155 & 308 & 1 & 1092 & 5 & MOTION \\
    Herring & 64 & 64 & 1 & 512 & 2 & IMAGE \\
    HouseTwenty & 40 & 119 & 1 & 2000 & 2 & DEVICE \\
    InlineSkate & 100 & 550 & 1 & 1882 & 7 & MOTION \\
    InsectEPGRegularTrain & 62 & 249 & 1 & 601 & 3 & EPG \\
    InsectEPGSmallTrain & 17 & 249 & 1 & 601 & 3 & EPG \\
    ItalyPowerDemand & 67 & 1029 & 1 & 24 & 2 & SENSOR \\
    LargeKitchenAppliances & 375 & 375 & 1 & 720 & 3 & DEVICE \\
    Lightning2 & 60 & 61 & 1 & 637 & 2 & SENSOR \\
    Lightning7 & 70 & 73 & 1 & 319 & 7 & SENSOR \\
    Mallat & 55 & 2345 & 1 & 1024 & 8 & SIMULATED \\
    Meat & 60 & 60 & 1 & 448 & 3 & SPECTRO \\
    MedicalImages & 381 & 760 & 1 & 99 & 10 & IMAGE \\ 
    MelbournePedestrian & 1138 & 2319 & 1 & 24 & 10 & TRAFFIC \\
    MiddlePhalanxOutlineAgeGroup & 400 & 154 & 1 & 80 & 3 & IMAGE \\
    MiddlePhalanxOutlineCorrect & 600 & 291 & 1 & 80 & 2 & IMAGE \\
    MiddlePhalanxTW & 399 & 154 & 1 & 80 & 6 & IMAGE \\
    MixedShapesRegularTrain & 500 & 2425 & 1 & 1024 & 5 & IMAGE \\
    MixedShapesSmallTrain & 100 & 2425 & 1 & 1024 & 5 & IMAGE \\
    MoteStrain & 20 & 1252 & 1 & 84 & 2 & SENSOR \\
    NonInvasiveFetalECGThorax1 & 1800 & 1965 & 1 & 750 & 42 & ECG \\
    NonInvasiveFetalECGThorax2 & 1800 & 1965 & 1 & 750 & 42 & ECG \\
    OSULeaf & 200 & 242 & 1 & 427 & 6 & IMAGE \\
    OliveOil & 30 & 30 & 1 & 570 & 4 & SPECTRO \\
    PLAID & 537 & 537 & 1 & 300 & 11 & DEVICE \\
    PhalangesOutlinesCorrect & 1800 & 858 & 1 & 80 & 2 & IMAGE \\
    Phoneme & 214 & 1896 & 1 & 1024 & 39 & AUDIO \\
    PickupGestureWiimoteZ & 50 & 50 & 1 & 361 & 10 & HAR \\
    PigAirwayPressure & 104 & 208 & 1 & 2000 & 52 & HEMODYNAMICS \\
    PigArtPressure & 104 & 208 & 1 & 2000 & 52 & HEMODYNAMICS \\
    PigCVP & 104 & 208 & 1 & 2000 & 52 & HEMODYNAMICS \\
    Plane & 105 & 105 & 1 & 144 & 7 & SENSOR \\
    PowerCons & 180 & 180 & 1 & 144 & 2 & DEVICE \\
    ProximalPhalanxOutlineAgeGroup & 400 & 205 & 1 & 80 & 3 & IMAGE \\
    ProximalPhalanxOutlineCorrect & 600 & 291 & 1 & 80 & 2 & IMAGE \\
    ProximalPhalanxTW & 400 & 205 & 1 & 80 & 6 & IMAGE \\
    RefrigerationDevices & 375 & 375 & 1 & 720 & 3 & DEVICE \\
    Rock & 20 & 50 & 1 & 2844 & 4 & SPECTRO \\
    ScreenType & 375 & 375 & 1 & 720 & 3 & DEVICE \\
    SemgHandGenderCh2 & 300 & 600 & 1 & 1500 & 2 & SPECTRO \\
    SemgHandMovementCh2 & 450 & 450 & 1 & 1500 & 6 & SPECTRO \\
    SemgHandSubjectCh2 & 450 & 450 & 1 & 1500 & 5 & SPECTRO \\
    ShakeGestureWiimoteZ & 50 & 50 & 1 & 385 & 10 & HAR \\
    ShapeletSim & 20 & 180 & 1 & 500 & 2 & SIMULATED \\
    ShapesAll & 600 & 600 & 1 & 512 & 60 & IMAGE \\
    SmallKitchenAppliances & 375 & 375 & 1 & 720 & 3 & DEVICE \\
    SmoothSubspace & 150 & 150 & 1 & 15 & 3 & SIMULATED \\
    SonyAIBORobotSurface1 & 20 & 601 & 1 & 70 & 2 & SENSOR \\
    SonyAIBORobotSurface2 & 27 & 953 & 1 & 65 & 2 & SENSOR \\
    StarLightCurves & 1000 & 8236 & 1 & 1024 & 3 & SENSOR \\
    Strawberry & 613 & 370 & 1 & 235 & 2 & SPECTRO \\
    SwedishLeaf & 500 & 625 & 1 & 128 & 15 & IMAGE \\
    Symbols & 25 & 995 & 1 & 398 & 6 & IMAGE \\
    SyntheticControl & 300 & 300 & 1 & 60 & 6 & SIMULATED \\
    ToeSegmentation1 & 40 & 228 & 1 & 277 & 2 & MOTION \\
    ToeSegmentation2 & 36 & 130 & 1 & 343 & 2 & MOTION \\
    Trace & 100 & 100 & 1 & 275 & 4 & SENSOR \\
    TwoLeadECG & 23 & 1139 & 1 & 82 & 2 & ECG \\
    TwoPatterns & 1000 & 4000 & 1 & 128 & 4 & SIMULATED \\
    UMD & 36 & 144 & 1 & 150 & 3 & SIMULATED \\
    UWaveGestureLibraryAll & 896 & 3582 & 1 & 945 & 8 & HAR \\
    UWaveGestureLibraryX & 896 & 3582 & 1 & 315 & 8 & HAR \\
    UWaveGestureLibraryY & 896 & 3582 & 1 & 315 & 8 & HAR \\
    UWaveGestureLibraryZ & 896 & 3582 & 1 & 315 & 8 & HAR \\
    Wafer & 1000 & 6164 & 1 & 152 & 2 & SENSOR \\
    Wine & 57 & 54 & 1 & 234 & 2 & SPECTRO \\
    WordSynonyms & 267 & 638 & 1 & 270 & 25 & IMAGE \\
    Worms & 181 & 77 & 1 & 900 & 5 & MOTION \\
    WormsTwoClass & 181 & 77 & 1 & 900 & 2 & MOTION \\
    Yoga & 300 & 3000 & 1 & 426 & 2 & IMAGE \\
    \bottomrule
    \end{tabularx}
\end{table*}

\begin{table*}[ht]
    \centering\scriptsize
    \caption{UEA datasets used as \ac{id} in our experiments. The table reports dataset statistics along with a \textit{Type} label, which is used to define near- and far-\ac{ood} splits in our benchmark.}
    \label{tab:uea_datasets}
    \begin{tabularx}{\columnwidth}{@{} >{\hsize=3cm}X CCCCCC @{}}
    \toprule
    \textbf{Dataset} & \textbf{TrainSize} & \textbf{TestSize} & \textbf{NumDimensions} & \textbf{SeriesLength} & \textbf{NumClasses} & \textbf{Type} \\
    \midrule
        ArticularyWordRecognition & 275 & 300 & 9 & 144 & 25 & MOTION \\
        AtrialFibrillation & 15 & 15 & 2 & 640 & 3 & ECG \\
        BasicMotions & 40 & 40 & 6 & 100 & 4 & HAR \\
        CharacterTrajectories & 1422 & 1436 & 3 & 119 & 20 & MOTION \\
        Cricket & 108 & 72 & 6 & 1197 & 12 & HAR \\
        DuckDuckGeese & 50 & 50 & 1345 & 270 & 5 & AUDIO \\
        ERing & 30 & 270 & 4 & 65 & 6 & HAR \\
        EigenWorms & 128 & 131 & 6 & 17984 & 5 & MOTION \\
        Epilepsy & 137 & 138 & 3 & 206 & 4 & HAR \\
        EthanolConcentration & 261 & 263 & 3 & 1751 & 4 & SPECTRO \\
        FaceDetection & 5890 & 3524 & 144 & 62 & 2 & EEG \\
        FingerMovements & 316 & 100 & 28 & 50 & 2 & EEG \\
        HandMovementDirection & 160 & 74 & 10 & 400 & 4 & EEG \\
        Handwriting & 150 & 850 & 3 & 152 & 26 & HAR \\
        Heartbeat & 204 & 205 & 61 & 405 & 2 & AUDIO \\
        InsectWingbeat & 25000 & 25000 & 200 & 20 & 10 & AUDIO \\
        JapaneseVowels & 270 & 370 & 12 & 25 & 9 & AUDIO \\
        LSST & 2459 & 2466 & 6 & 36 & 14 & OTHER \\
        Libras & 180 & 180 & 2 & 45 & 15 & HAR \\
        MotorImagery & 278 & 100 & 64 & 3000 & 2 & EEG \\
        NATOPS & 180 & 180 & 24 & 51 & 6 & HAR \\
        PEMS-SF & 267 & 173 & 963 & 144 & 7 & OTHER \\
        PenDigits & 7494 & 3498 & 2 & 8 & 10 & MOTION \\
        PhonemeSpectra & 3315 & 3353 & 11 & 217 & 39 & AUDIO \\
        RacketSports & 151 & 152 & 6 & 30 & 4 & HAR \\
        SelfRegulationSCP1 & 268 & 293 & 6 & 896 & 2 & EEG \\
        SelfRegulationSCP2 & 200 & 180 & 7 & 1152 & 2 & EEG \\
        SpokenArabicDigits & 6599 & 2199 & 13 & 65 & 10 & SPEECH \\
        StandWalkJump & 12 & 15 & 4 & 2500 & 3 & ECG \\
        UWaveGestureLibrary & 120 & 320 & 3 & 315 & 8 & HAR \\
    \bottomrule
    \end{tabularx}
\end{table*}

\section{Modality Remapping for Singleton Types}\label{app:modality_remap}
Some dataset types appear only once in the UCR/UEA archives, which makes it ambiguous to define near-shift counterparts. To ensure that all datasets participate meaningfully in both near- and far-shift evaluations, we follow a fixed modality-to-most-similar modality mapping. This mapping is chosen based on signal similarity, preprocessing pipelines. Tables~\ref{tab:uea_modality_map} and~\ref{tab:ucr_modality_map} provide the mapping rules and their rationale for UCR and UEA, respectively. The corresponding dictionaries are released with our code to ensure reproducibility.

\begin{table*}[ht]
\centering\scriptsize
\caption{UCR Modality-to-Closest Modality Map with explanations.}
\label{tab:ucr_modality_map}
\begin{tabularx}{\columnwidth}{@{} >{\hsize=2cm}X  >{\hsize=2cm}X X @{}}
\toprule
\textbf{Modality} & \textbf{Closest} & \textbf{Rationale} \\
\midrule
HAR & MOTION & Both describe human activity from body-worn sensors. \\
MOTION & HAR & Reciprocal mapping. \\
DEVICE & SENSOR & Device telemetry is closest to generic sensor data. \\
SENSOR & DEVICE & Reciprocal mapping. \\
IMAGE & SPECTRO & Image-derived series resemble spectrogram-like representations. \\
SPECTRO & AUDIO & Spectrograms are transformations of audio. \\
AUDIO & SPECTRO & Reciprocal mapping. \\
ECG & EOG & Both are biomedical electrophysiological signals. \\
EOG & ECG & Reciprocal mapping. \\
EPG & ECG & Similar biomedical electrophysiology tasks. \\
HEMODYNAMICS & ECG & Both are physiological signals often co-recorded. \\
TRAFFIC & SENSOR & Traffic counts are sensor-based temporal signals. \\
SIMULATED & SENSOR & Simulations mimic sensor dynamics. \\
OTHER & SPECTRO & Default mapping to a general-purpose representation. \\
\bottomrule
\end{tabularx}
\end{table*}

\begin{table*}[ht]
\centering\scriptsize
\caption{UEA Modality-to-Closest Modality Map with explanations.}
\label{tab:uea_modality_map}
\begin{tabularx}{\columnwidth}{@{} >{\hsize=2cm}X  >{\hsize=2cm}X X @{}}
\toprule
\textbf{Modality} & \textbf{Closest} & \textbf{Rationale} \\
\midrule
MOTION & HAR & Both capture body dynamics via wearable sensors. \\
HAR & MOTION & Reciprocal mapping; same rationale. \\
EEG & ECG & Both are biomedical electrophysiological signals with low SNR. \\
ECG & EEG & Reciprocal mapping. \\
AUDIO & SPEECH & Speech is a subset of acoustic audio, often using MFCC/spectrogram features. \\
SPEECH & AUDIO & Reciprocal mapping. \\
SPECTRO & AUDIO & Spectrograms originate from audio preprocessing. \\
OTHER & SPECTRO & Default mapping; spectrogram is a general representation for heterogeneous signals. \\
\bottomrule
\end{tabularx}
\end{table*}

\end{document}